# An Estimation Method of Measuring Image Quality for Compressed Images of Human Face

Abhishek Bhattacharya
Department of Computer Science, Institute of Engineering & Management

Tanusree Chatterjee
Department of Computer Science, Regent Education and Research Foundation

***Abstract-*** *Nowadays digital image compression and decompression techniques are very much important. So our aim is to calculate the quality of face and other regions of the compressed image with respect to the original image. Image segmentation is typically used to locate objects and boundaries (lines, curves etc.)in images. After segmentation the image is changed into something which is more meaningful to analyze. Using Universal Image Quality Index(Q),Structural Similarity Index(SSIM) and Gradient-based Structural Similarity Index(G-SSIM) it can be shown that face region is less compressed than any other region of the image.*

***Keywords – Image analysis, Segmentation, Image quality, Image region analysis***

## I. Introduction

The task of recognizing a person's face from an frontal image is very easy to human eyes though it is difficult for a machine to perform the same task. In fact it's a very wide study to perfectly detect a human face (skin) which is very much difficult for a machine. Now there are several types of color spaces like RGB, YCrCb, HSV etc. Now in this paper we use YCrCb color space for two reasons-First, The chrominance information for modeling human skin color can be achieved this color space. Second, for all kind of images, coding is used in YCrCb color format.

Now we know that Cr and Cb are the blue and red difference of chroma component. It is proved by experiments that the range of skin color of a human being is always between 133-173 and 77-127 respectively. Now the darkness and fairness of human skin is always dependent on 'Y' which is the luma component of the color image .Actually 'Y' is the brightness of an image.

Now after recognizing the face of images especially for video calling purpose or live transmission of Index(Q),Structural Similarity Index(SSIM) and Gradient-based Structural Similarity Index(G-SSIM)news reading images, compression is very much required to get the fast transmission. When an original image file is compressed in other web images format like FLV,3GP,MP4,MPEG and DIVX, the compression techniques compress the images of the video files. Our objective is to show that in all the images compression techniques the face region is less compressed according to the other regions of the image. Visually human eyes can't detect compression ratio of the face and the other regions of the image. So to prove this mathematically we have used Universal Image Quality Index (Q), Structural Similarity Index (SSIM) and Gradient-based Structural Similarity Index (G-SSIM).

## II. RELATED WORK

A number of studies have demonostrated the factors of quality of images after compression.Sheet el al. have worked on Universal Image Quality Index (Q), Structural Similarity Index (SSIM) and Gradient-based Structural Similarity Index(G-SSIM) [1].Bhattacharya et al. have worked on images segmentation related with this context[2]. Chatterjee et al. worked on segmentation of gray image using projection [3].The basic idea of Universal Image Quality Index (Q), Structural Similarity Index (SSIM) and Gradient-based Structural Similarity Index (G-SSIM) is inspired from the structural model[1].Sheet et al. [1] have shown that all the applications are applied on face and skin region of image to make analysis more prominent. The universal image quality index is used as image quality distortion measure. It in mathematically defined by modeling the image distortion relative to the reference image as a combination of three factors: loss of correlation, luminance distortion and contrast distortion [1].The structural similarity index is a method for measuring the similarity between two images. The SSIM index is a full reference metric which is based on an initial uncompressed image as reference. SSIM is designed to improve traditional metrics such as signal-to-noise ratio and mean square error (MSE) which has proved to be inconsistent with human eye perception [1].SSIM often fails to measure badly blurred images. Based on this observation, a improved method GSSIM can be applied to reference image. Experimental results show that GSSIM is more consistent on HVS image then SSIM and PSNR especially for blurred images. Chai et al. worked on face segmentation with skin color map [4].The idea behind this method is to find out a fast, reliable and effective algorithm which exploits the spatial distribution characteristics of human skin color. A universal skin-color map was derived and used on the chrominance component of the input image to detect pixels with skin-color appearance. Experimental results demonstrated that individual color features can be defined by the presence of Cr values from 136 to 156 and Cb values from 110 to123[4].

## III. PROCEDURAL DETAILS

Image compression is all about compressing the data of image so that it can be transmitted more easily.In a particular image there are several areas where color information is the dsame.On the other hand human eye can not distinghish a very wide range of different colors.So one can use a computer program to compress an image by removing the redundant informatons from it.The universal image quality index(Q) models the distortion in an image using three factors: loss of correlation,luminance distortion and contrast distortion.Since image is often spatially varying,it is approproate to measure statistical features locally and then combine them.This method applies a sliding window method throughout the amage for assessing total image quality.

If two images f and g are considered as a matrices with M column and N rows containing pixel values f[i,j], g[i,j], respectively ($0 \geq i > M$, $0 \geq j > N$), the universal image quality index Q may be calculated as a product of three components:

$$Q = \frac{\sigma_{fg}}{\sigma_f \sigma_g} \cdot \frac{2\bar{f}\bar{g}}{(\bar{f})^2 + (\bar{g})^2} \cdot \frac{2\sigma_f\sigma_g}{\sigma_f^2 + \sigma_g^2}$$

where

$$\bar{f} = \frac{1}{MN}\sum_{i=0}^{M-1}\sum_{j=0}^{N-1} f[i,j] \qquad \bar{g} = \frac{1}{MN}\sum_{i=0}^{M-1}\sum_{j=0}^{N-1} g[i,j]$$

$$\sigma_{fg} = \frac{1}{M+N-1}\sum_{i=0}^{M-1}\sum_{j=0}^{N-1}(f[i,j]-\bar{f})(g[i,j]-\bar{g})$$

$$\sigma_f^2 = \frac{1}{M+N-1}\sum_{i=0}^{M-1}\sum_{j=0}^{N-1}(f[i,j]-\bar{f})^2 \qquad \sigma_g^2 = \frac{1}{M+N-1}\sum_{i=0}^{M-1}\sum_{j=0}^{N-1}(g[i,j]-\bar{g})^2$$

The first component is the correlation coefficient, which measures the degree of linear correlation between images *f* and *g*. It varies in the range [-1, 1]. The best value 1 is obtained when *f* and *g* are linearly related, which means that $g[i,j] = af[i,j]+b$ for all possible values of *i* and *j*. The second component, with a value range of [0,1], measures how close the mean luminance is between images. Since $\sigma_f$ and $\sigma_g$ are considered as estimates of the contrast of *f* and *g*, the third component measures how similar the contrasts of the images are. The value range for this component is also [0, 1].The range of values for the index *Q* is [-1, 1]. The best value 1 is achieved if and only if the images are identical. The structural similarity (SSIM) index is a method for measuring the similarity between two images. In other words, the measuring of image quality based on an initial uncompressed or distortion-free image as reference. SSIM is designed to improve on traditional methods like peak signal-to-noise ratio (PSNR) which have proven to be inconsistent with human eye perception.

The difference with respect to other techniques mentioned previously such PSNR is that these approaches estimate perceived errors; on the other hand, SSIM considers image degradation as perceived change in structural information. Structural information is the idea that the pixels have strong inter-dependencies especially when they are spatially close.

The SSIM metric is calculated on various windows of an image. The measure between two windows $x$ and $y$ of common size N×N is:

$$\mathrm{SSIM}(x,y) = \frac{(2\mu_x\mu_y + c_1)(2\sigma_{xy} + c_2)}{(\mu_x^2 + \mu_y^2 + c_1)(\sigma_x^2 + \sigma_y^2 + c_2)}$$

Where

$\mu_x$ the average of $x$;

$\mu_y$ the average of $y$;

$\sigma_x^2$ the variance of $x$;

$\sigma_y^2$ the variance of $y$;

$\sigma_{xy}$ the covariance of $x$ and $y$;

$c_1=(k_1L)^2$, $c_2=(k_2L)^2$ two variables to stabilize the division with weak denominator;

$L$ the dynamic range of the pixel-values (typically this is $2^{\#bits\ per\ pixel}-1$);

$k_1=0.01$ and $k_2=0.03$ by default

In order to lower the complexity, we substitute a gradient value in Fast SSIM. While images of real-world scenes vary greatly in their absolute luma and chroma distributions, the gradient magnitudes of natural images generally obey heavy tailed distribution laws. Indeed, some no-reference image quality assessment algorithms use the gradient image to assess blur severity. Similarly, the performance of the Gradient-based SSIM index suggests that applying SSIM on the gradient magnitude may yield higher performance. The gradient is certainly responsive to image variation This approximation is based upon a simple expansion of the gradient. The contrast c(x, y) and structure s(x, y) terms of the Fast SSIM index algorithm are then defined[5]:

$$c(x,y) = \frac{(2\mu_{Gx}\mu_{Gy} + C_2)}{(\mu_{Gx}^2 + \mu_{Gy}^2 + C_2)}$$

$$s(x,y) = \frac{(\mu_{GxGy} + C_3)}{(\mu_{Gx}\mu_{Gy} + C_3)}$$

Where $C_3=C_2/2$ and

$$\mu_{Gx} = \frac{1}{N}\sum_{i=1}^{N}|\nabla x_i|$$

$$\mu_{GxGy} = \frac{1}{N}\sum_{i=1}^{N}|\nabla x_i||\nabla y_i|$$

## IV. EXPERIMENTAL RESULTS

We have implemented the method using Matlab7.1 on multiple images and we have received quite satisfactory results. However a few outputs are shown here. The images taken were in TIFF format. All experiments were conducted on a Intel Core2Duo 2.2 GHz platform. Table 1 makes it clear that the for face region SSIM,GSSIM and Q produce higher values than body regions. So it can be concluded that face region is less compressed than any other region of the image. A graphical comparison of the same are shown in fig(ii). In Figure (i) (a) the original reference image is shown. In Figure(i) (b) compressed image(10% ) quality, In Figure(i) (c) compressed image(20% ) quality ,In Figure(i) (d) compressed image(30% ) quality, In Figure(i) (e) compressed image(40% ) quality, In Figure(i) (f) compressed image(50% ) quality, In Figure(i) (g) compressed image(60% ) quality, In Figure(i) (h) compressed image(70% ) quality, In Figure(i) (i) compressed image(80% ) quality, In Figure(i) (j) compressed image(90% ) quality, In Figure(i) (k) compressed image(100% ) quality is shown. In Figure (i) (m) and (n) the face region and body regions are shown respectively.

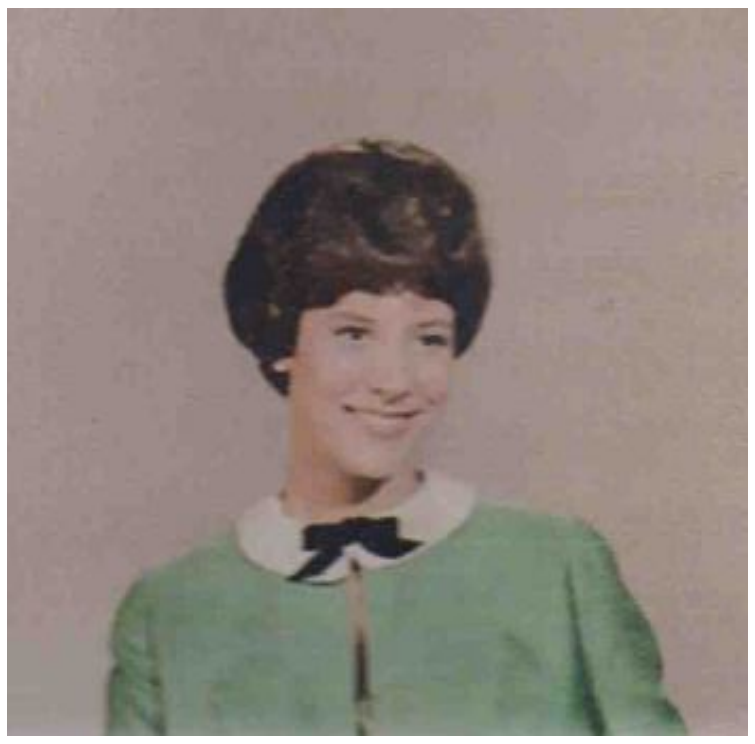

(a)

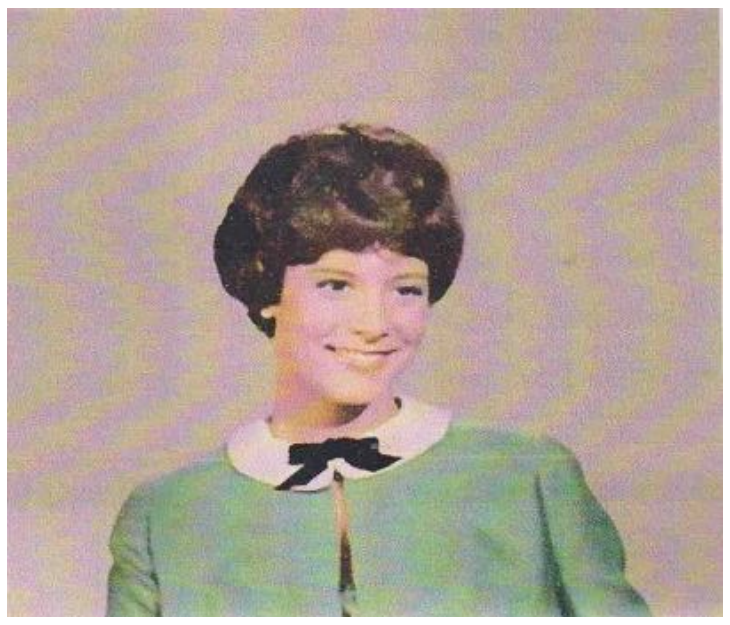

(b)

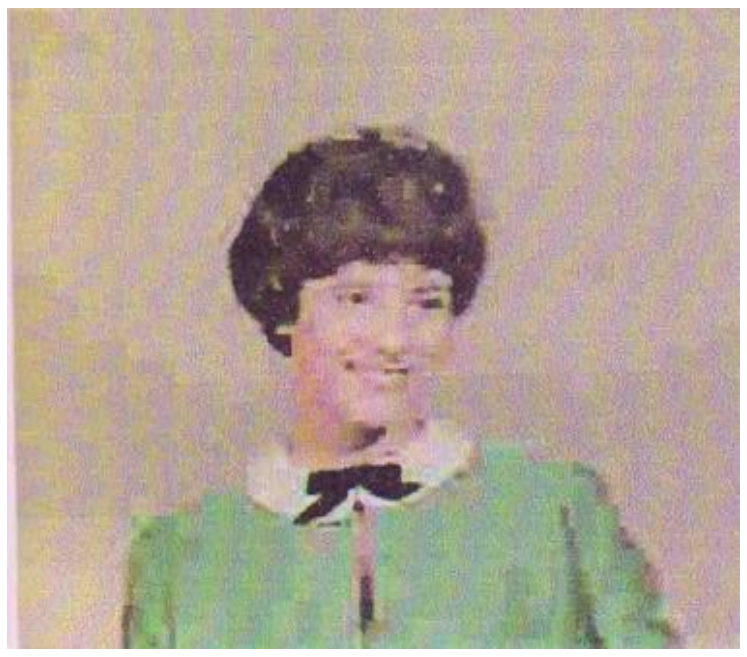

(c)

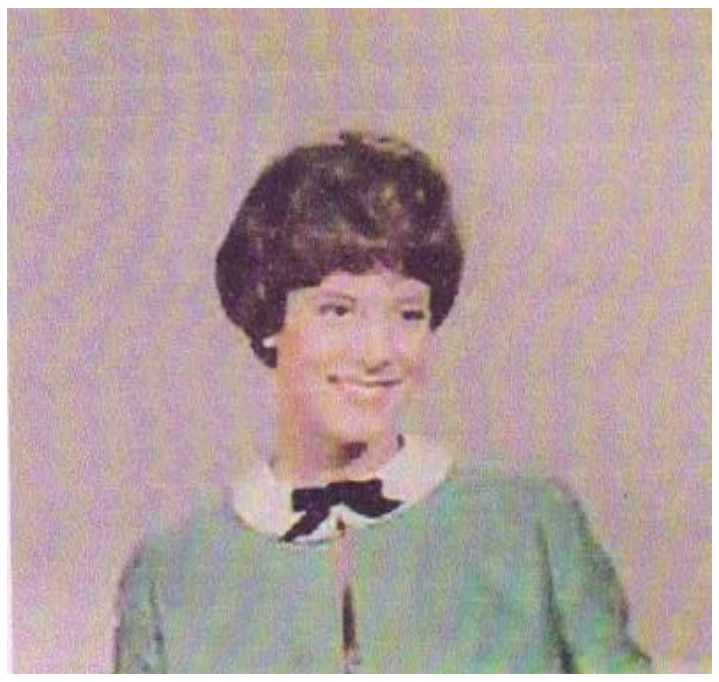

(d)

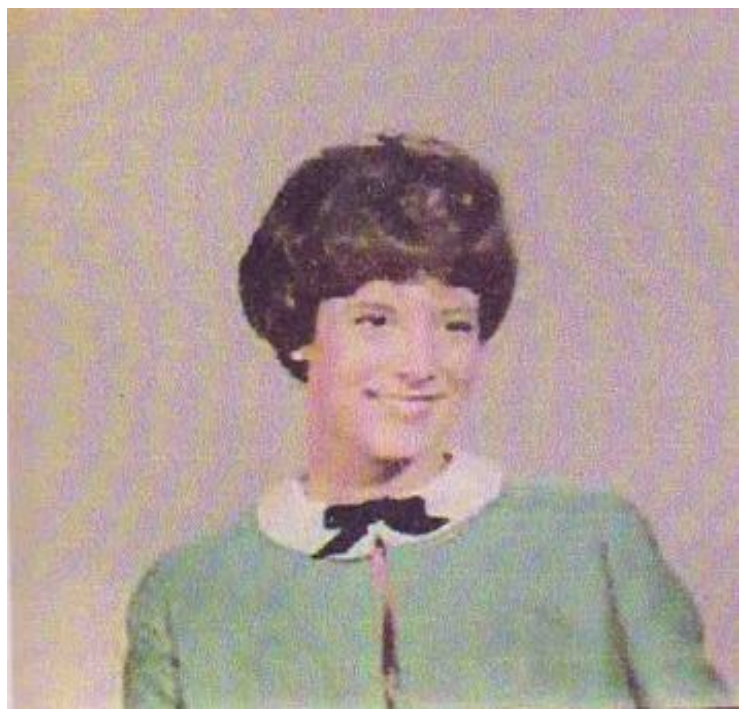

(e)

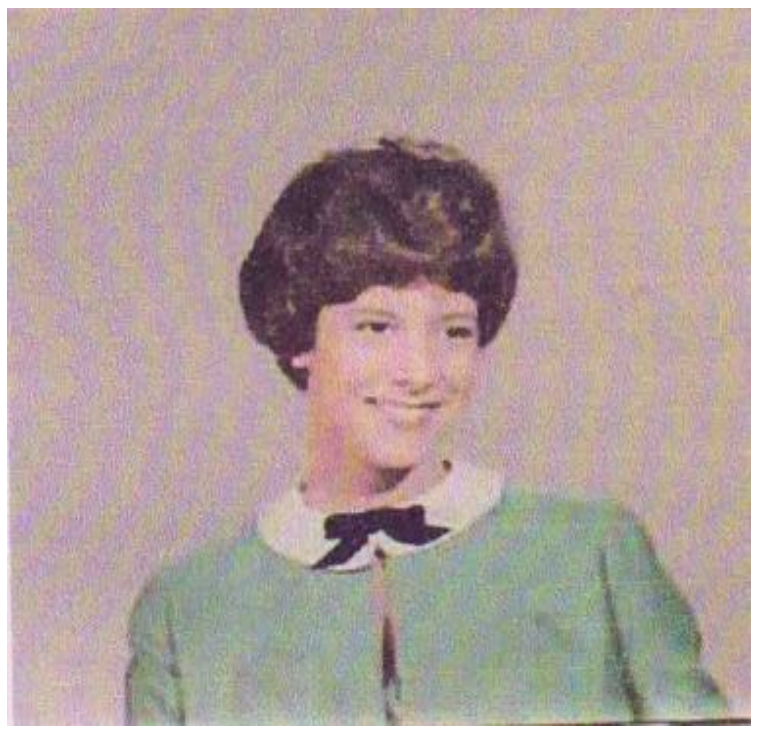

(f)

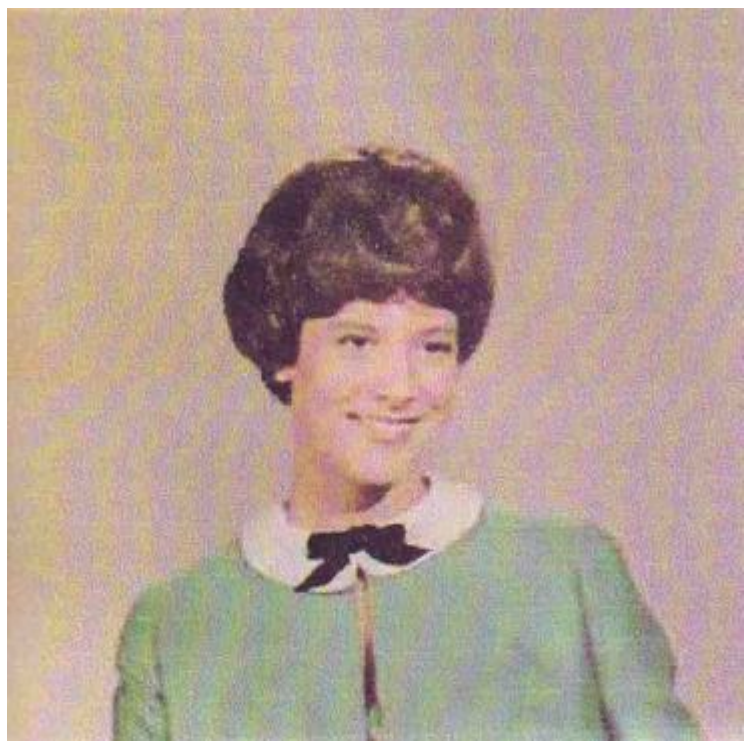

(g)

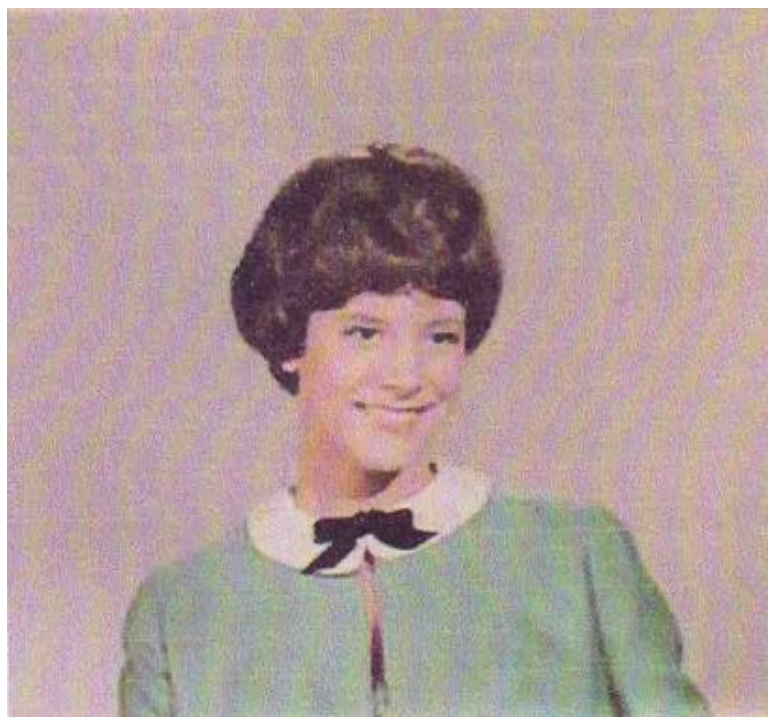

(h)

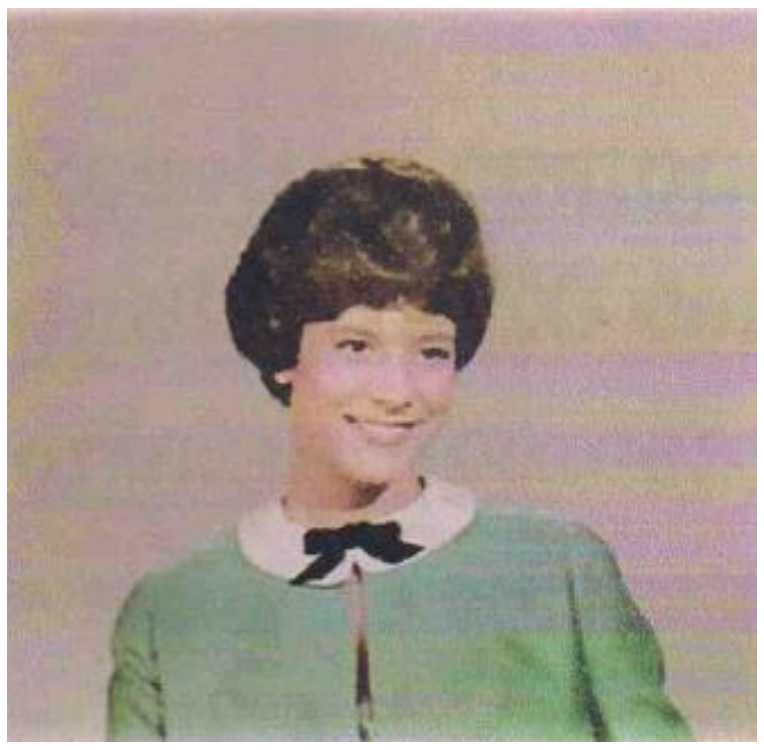

(i)

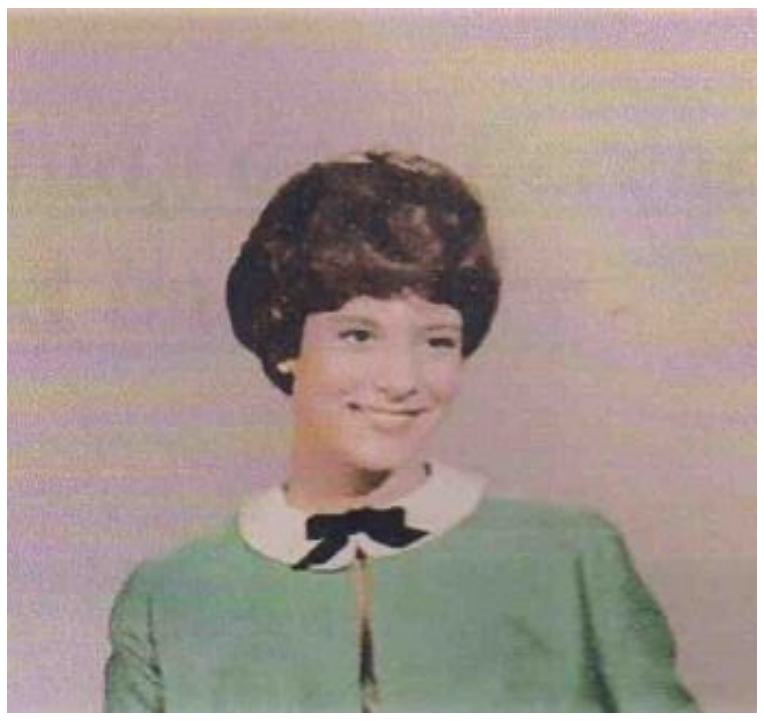

**(j)**

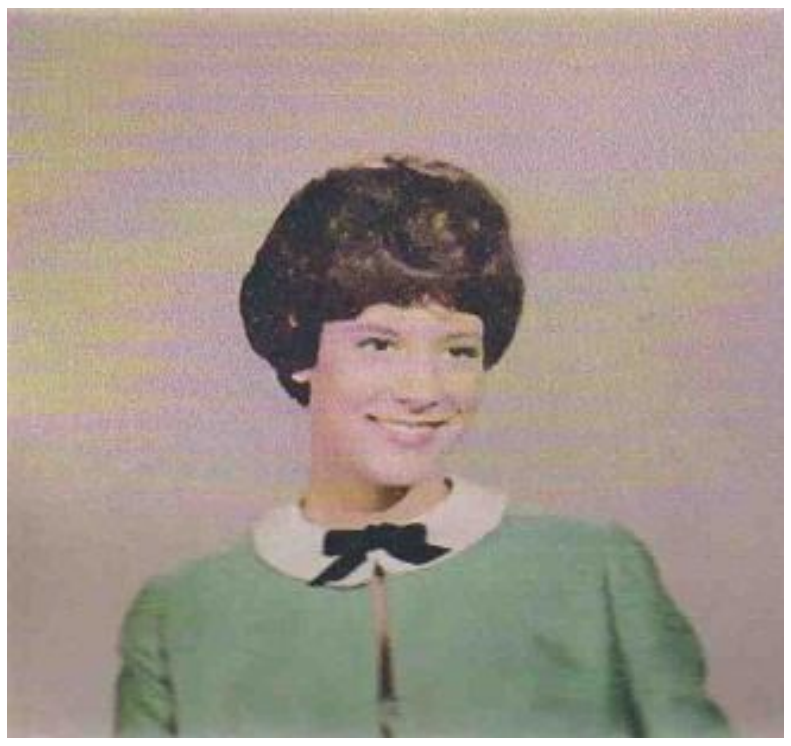

**(k)**

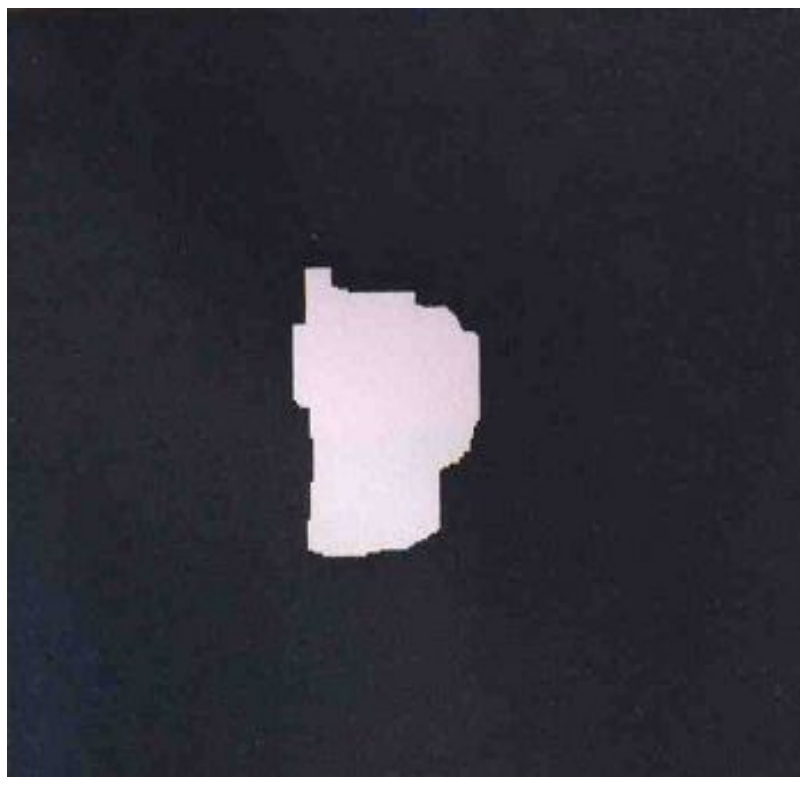

**(n)**

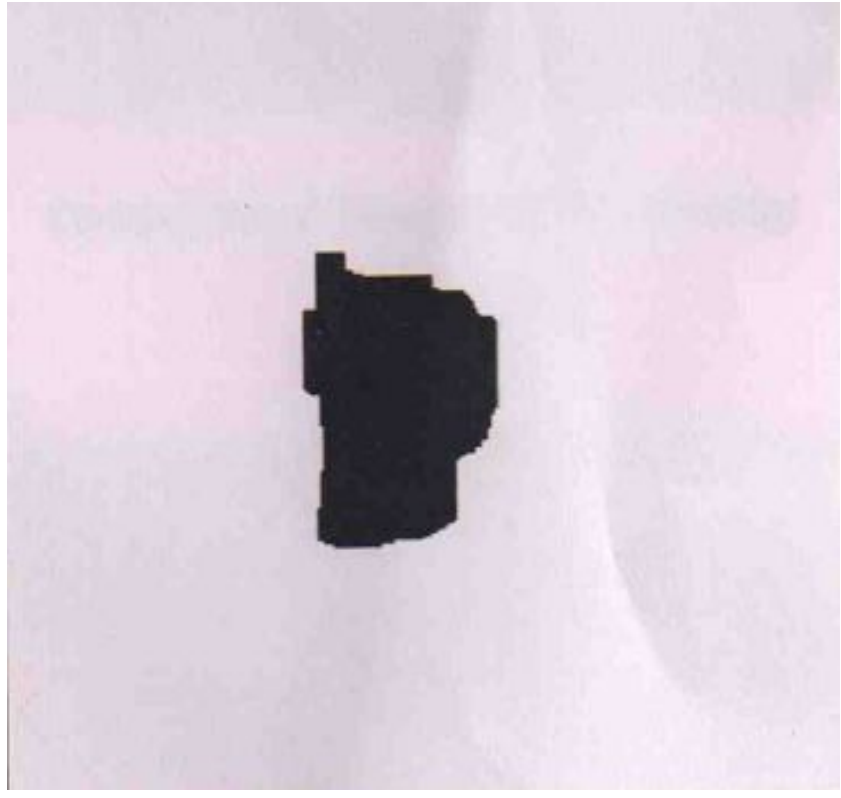

**(o)**

| Face Quality | Body Quality | Face SSIM | Body SSIM | Face GSSIM | Body GSSIM |
|---|---|---|---|---|---|
| 0.735958299 | 0.619265678 | 0.820405841 | 0.769979111 | 0.999226036 | 0.999568373 |
| 0.829313751 | 0.624684403 | 0.899929133 | 0.814301773 | 0.999766737 | 0.999881885 |
| 0.870841760 | 0.654425948 | 0.926339864 | 0.848026567 | 0.999884378 | 0.999943486 |
| 0.888676177 | 0.632148493 | 0.938727995 | 0.855834163 | 0.999921523 | 0.999960910 |
| 0.903363645 | 0.606649817 | 0.947728780 | 0.866117072 | 0.999941406 | 0.999975812 |
| 0.912620523 | 0.592910206 | 0.954184915 | 0.872629536 | 0.999962419 | 0.999981224 |
| 0.930153935 | 0.564726399 | 0.963563824 | 0.882819885 | 0.999976403 | 0.999986809 |
| 0.943169910 | 0.529969600 | 0.971089965 | 0.895699355 | 0.999987087 | 0.999993079 |
| 0.966637365 | 0.585064240 | 0.983437933 | 0.915191921 | 0.999995819 | 0.999997550 |
| 0.999046805 | 0.977693185 | 0.999636261 | 0.994544491 | 0.999999737 | 0.999999811 |

Figure(i)

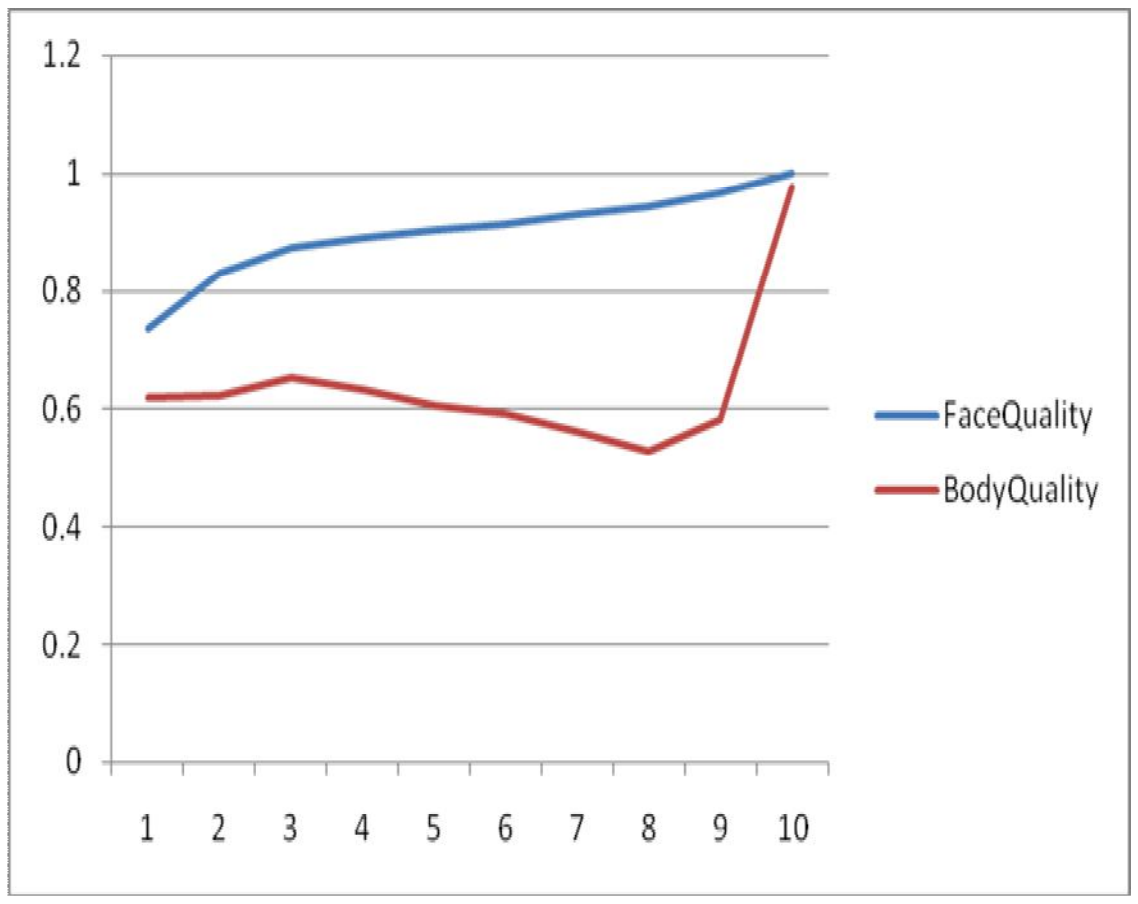

(a)

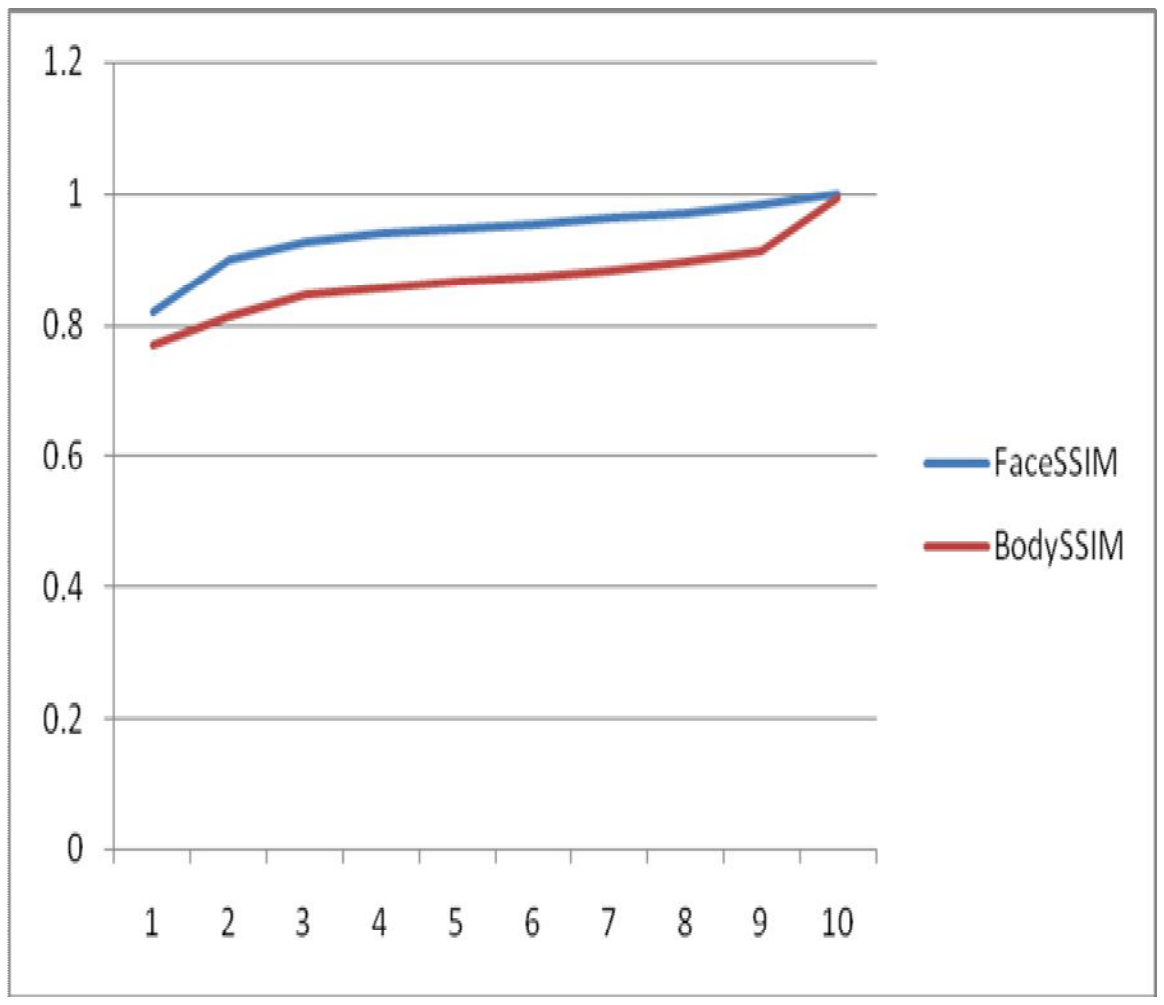

(b)

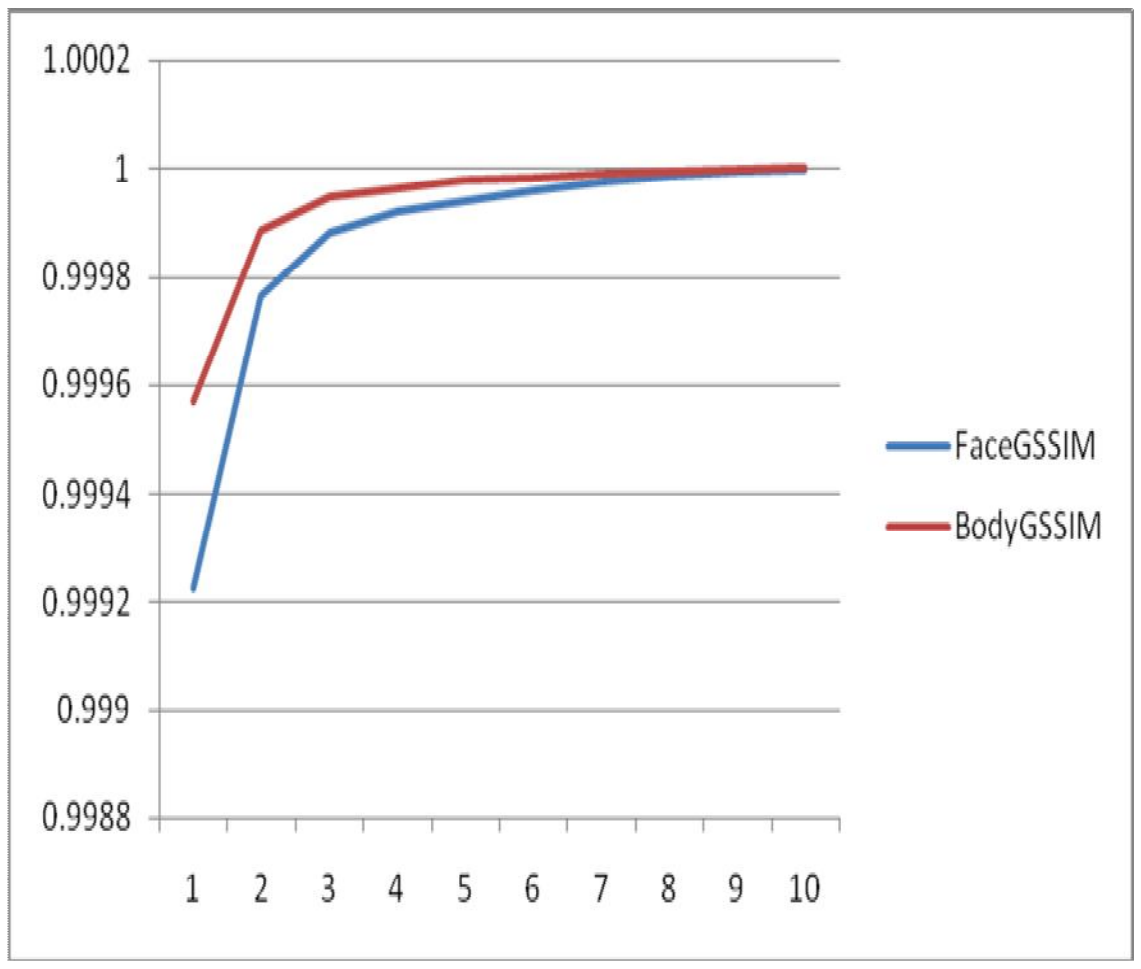

(c)
Figure (ii)

## V. CONCLUSION AND FUTURE SCOPE

From the above experiments on JPEG image compression we can reach in the decision that for any lossy and lossless image compression the face region is compressed at a very low rate where as the other regions of the image are compressed at a very high rate. As the face of the person is very sensitive at the time of communication, thus the image quality of the face region have to be reserved properly .At the time of image compression the data bits are lost which is the cause of degradation of the image quality.

From the above we can also conclude that at the time of compressing an image we should always convert RGB or HSV image into YCrCb format. This format is much more effective than the other format for compression purpose. There are two different purposes of using YCrCb format. First an effective use of chrominance information for medaling human skin color can be achieved in this color space. Second, this format is typically used in image compression, and therefore the use of the same, instead of another format for segmentation will avoid the extra computation required in conversion.

In this paper the comparative study has been done on JPEG image which has a tremendous scope of future application if handled properly with a supporting video. Due to lack of proper images and video the video compression could not be applied properly on video compression techniques. For different percentage of compression on JPEG image, it has been proved that the face region is compressed very less according to the other regions of the image. So at the time of video translation it will require less bandwidth and less storage space. The proper application of this will be helpful for video chatting, video calling and the sector of film editing also.

## AUTHORS PROFILE

**First Author** –Abhishek Bhattacharya, M.Tech (CSE) from BIT,Mesra, Assistant Professor in Department of Computer Science, Institute of Engineering & Management, Saltlake, Kolkata.

**Second Author** -Tanusree Chatterjee, M.Tech (CSE) from WBUT, Assistant Professor in Department of Computer Science, Regent Education and Research Foundation,